\documentclass[10pt,twocolumn,letterpaper]{article}

\usepackage[table, svgnames]{xcolor}
\usepackage[utf8]{inputenc}
\usepackage{lmodern}
\usepackage{tikz}
\usepackage{iccv}
\usepackage{lmodern}
\usepackage[french,german,english]{babel}
\usepackage{xparse}
\usepackage{times}
\usepackage{epsfig}
\usepackage{graphicx}
\usepackage{amsmath}
\usepackage{amssymb}
\usepackage{siunitx}
\usepackage{tabularx,booktabs}
\usepackage[
    activate=true,
    tracking=true,
    kerning=true,
    spacing=true,
    factor=1100,
    stretch=10,
    shrink=10]{microtype}

\usepackage{mathtools}
\usepackage{enumitem}
\usepackage[xindy, acronym]{glossaries}
\usepackage{calc}
\usepackage{csquotes}
\usepackage{subcaption}

\makeglossaries
\newacronym{coco}{COCO}{Common Objects in COntext}
\newacronym{muse}{MUSE}{Multilingual Unsupervised or Supervised word Embeddings}
\newacronym{rnn}{RNN}{Recurrent Neural Networks}
\newacronym{cnn}{CNN}{Convolutional Neural Networks}
\newacronym{smt}{SMT}{Statistical Machine Translation}
\newacronym{nmt}{NMT}{Neural Machine Translation}
\newacronym{w2v}{W2V}{Word2Vec}
\newacronym{cbow}{CBOW}{Continuous Bag-of-Words}
\newacronym{sru}{SRU}{Simple Recurrent Unit}
\newacronym{bv}{BIVEC}{Bilingual Word Representations}
\newacronym{dsve}{DSVE}{Deep Semantic-Visual Embedding}
\newacronym{vse}{VSE++}{Visual-Semantic Embedding}

\setlist[itemize]{label={\tiny\raisebox{1ex}{\textbullet}}}

\usepackage[pagebackref=true,breaklinks=true,letterpaper=true,colorlinks,bookmarks=false]{hyperref}

\iccvfinalcopy 

\def\iccvPaperID{3046} 

\ificcvfinal\fi

\definecolor{visualpath}{HTML}{cc9cf9}
\definecolor{textpath}{HTML}{72b3b3}

\begin{document}

\title{Image search using multilingual texts: a cross-modal learning approach between image and text}

\author{Maxime Portaz\\
Qwant Research\\
{\tt\small m.portaz@qwant.com}
\and
Hicham Randrianarivo\\
Qwant Research\\
{\tt\small h.randrianarivo@qwant.com}
\and
Adrien Nivaggioli\\
Qwant Research\\
{\tt\small a.nivaggioli@qwant.com}
\and
Estelle Maudet\\
Qwant Research\\
{\tt\small e.maudet@qwant.com}
\and
Christophe Servan\\
Qwant Research\\
{\tt\small c.servan@qwant.com}
\and
Sylvain Peyronnet\\
Qwant Research\\
{\tt\small s.peyronnet@qwant.com}
}

\maketitle

\begin{abstract}

Multilingual (or cross-lingual) embeddings represent several languages in a unique vector space. Using a common embedding space enables for a shared semantic between words from different languages.
In this paper, we propose to embed images and texts into a unique distributional vector space, enabling to search images
by using text queries expressing information needs related to the (visual) content of images, as well as using image similarity.
Our framework forces the representation of an image to be similar to the representation of the text that describes it. Moreover, by using multilingual embeddings we ensure that words from two different languages have close descriptors and thus are attached to similar images.
We provide experimental evidence of the efficiency of our approach by experimenting it on two datasets: \gls{coco}~\cite{Lin2014} and Multi30K~\cite{Elliott2016}.
\end{abstract}


\section{Introduction}
\label{sec:introduction}

\begin{figure}[t]
    \centering
    \includegraphics[width=\columnwidth]{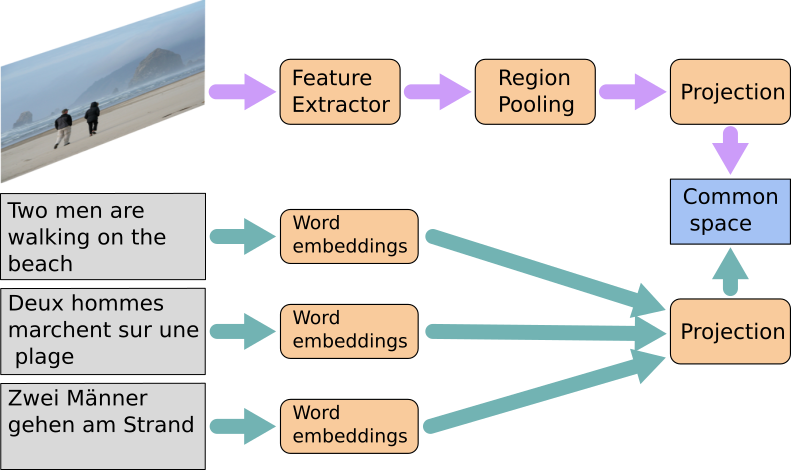}
    \caption{The Overall pipeline of our method. The method is decomposed into two paths. The visual path that extracts a representation from the image and the text path that extracts a representation from texts. The text path project texts of different languages in the same vector space.}
    \label{fig:overview}
\end{figure}


Deep Neural Networks have shown improvement in state-of-the-art in different tasks, such as image classification~\cite{KrizhevskyImageNetNetworks, He2016}, word embeddings~\cite{Luong2015}, text classification~\cite{SocherRecursiveTreebank} or image retrieval~\cite{Babenko2014NeuralRetrieval, SharifCNNRecognition}. 
They can be used to embed data into features vectors that were primarily used for information retrieval.
On texts, evolution of these embeddings (namely multilingual embeddings) made it possible to solve multilingual tasks such as cross-lingual classification of texts.
On images, they allow comparison based on image content, with text or image queries.

Cross-modal networks are using recurrent and convolutional networks together in order to embed texts and images in a common vector space.
This research area is heavily studied and we present a short related work in section~\ref{sub:related_works}.


While many approaches embed visual and semantic information together, they are most of the time limited to only one language.
In this paper, we propose a novel approach for multilingual text and image embeddings (section~\ref{sec:model}).
Our method uses \gls{cnn} to extract Image information and aligned multilingual Word embeddings and \gls{rnn} to produce a text representation.
With this framework, we provide image and text embeddings that can be trained to produce comparable features. We can thus retrieve images from text, and texts from an image, with a multilingual representation.

More precisely, we propose two approaches. The first uses \gls{bv} embeddings (see~\cite{Luong2015}) and improves on the state-of-the-art for English when trained on another language.
The second uses \gls{muse} (see~\cite{Conneau2018}). It enables recognition for several languages with only one model, with a slight performance drawback. More precisely, by using \gls{muse} aligned embeddings in 30 different languages, we can retrieve images with languages never seen by the model.

The major lesson learned with respect to our method is that it provides close to the state-of-the-art performance in English-only context, but enables the use of multilingual datasets to improve results.
We show in section~\ref{sec:xp} that adding a new language during the training phase improves image retrieval for other languages.
The experiments show
that the approach based on \gls{bv} embeddings gives a \SI{3.35}{\percent} increase in performance on the \gls{coco} dataset, and a \SI{15.15}{\percent} increase on Multi30K.
By using \gls{muse}, we are able to encode more languages in the same model. The experiments for that approach show that adding other languages gives a small decrease in performance for English, but increases the recall for a multilingual environment. Indeed, we obtain \SI{49.38}{\percent} recall@10 on Multi30K dataset for image retrieval from captions in 4 languages.

\section{Related Work}
\label{sub:related_works}


\subsection{Text embeddings}
The use of word representation is an important step when it comes to search for information from text documents.
In order to perform this task, we want to be able to extract meaningful embeddings from words.
One useful property of word embeddings is that words with a similar meaning must have representations with a close distance.
A lot of people work on this task but the most popular methods are \gls{w2v}~\cite{Mikolov2013} and FastText~\cite{Bojanowski2016}
These methods are simplified methods of the neural language model proposed by Bengio et al.~\cite{Bengio2003} with several tricks to boost performance.

\subsection{Multilingual word embeddings.}
Word embeddings can be used in multilingual tasks (e.g. machine translation or cross-lingual document classification) by training a model independently for each language.
However, the representations will be in separate vector spaces, which means that the same words in different languages will most likely have different representations.

There are several methods to solve this problem. One consists in training both models independently and then to learn a mapping from one representation to the other ~\cite{Faruqui2015,Mikolov2013}. It is also possible to constrain the training to keep the representations of similar words close to each other~\cite{Hermann2013,P2014}. Finally, the training can be performed jointly using parallel corpora~\cite{Klemetiev2012}.

In the latter category, the \gls{bv} approach~\cite{Luong2015}
tries to predict words based on the inner context of the sentence like \gls{w2v} does but also uses words in the source sentence to predict words in the target sentence (and conversely). Thus, for each update in \gls{w2v},  \gls{bv} performs 4 updates: source to source, source to target, target to target and target to source. This leads to a common representation for the two languages.


Recently, \gls{muse}~\cite{Conneau2018} proposes to learn a mapping from several word embeddings trained independently.
This approach enables mapping between word embeddings from different languages.

\subsection{Cross-modal representation}

In order to provide queries as sentences or as images, the image embeddings and the text embeddings must be comparable, i.e. in the same representation space.
Recent works have shown the possibility to learn text and image representation simultaneously~\cite{Karpathy2015,Faghri2018,Engilberge2018}.
They rely on cross-modal networks, that are able to extract information from images and read the caption describing it.
Those networks use word embeddings followed by a \gls{rnn} to encode sentence embeddings in the same space as the image embeddings, extracted with a \gls{cnn}.

\gls{cnn} methods provide ways to encode images in meaning full embeddings.
Prior works considered image similarity based on the categories~\cite{Guillaumin2009,Wang2014}.
Recent approaches~\cite{Faghri2018, Engilberge2018} use ResNet~\cite{Xie2017} as CNN image features extractor.
For the text part on the network, they use \gls{w2v} or Skip-Thought~\cite{Kiros2015,Engilberge2018} word embeddings, followed by a multi-layer \gls{rnn} to encode the sentence.

Those methods are multi-branches networks.
The loss function has to be a comparative loss, with a similarity function.
The similarity function is generally estimated using the Euclidean distance, or the cosine similarity.
To train network with this type of loss function, we use a Siamese network~\cite{Bromley1993,Bertinetto2016,Yi2014}.
Siamese networks have been extended to triplet networks with three branches in order to give better results~\cite{Hoffer2015,Schroff2015}.
Triplet is composed of an anchor image, an example of a similar image (positive image) and a dissimilar example (negative image).

Those methods enable to learn complex image representation with few examples, as it is possible to select the best triplet example for the training~\cite{Wang2014,Gordo2017a,Portaz2017}.
Our model is based on triplet networks, with each branch based on text or on images, interchangeably.




\section{Multilingual Joint Text Image Embedding}\label{sec:model}

We present a model for multilingual and cross-modal (joint learning of text and image representation) embeddings.
We propose to embed images and sentences from different languages in the same space $[-1 , 1 ]^d$, where the distance between two elements (image or sentence) is inversely proportional to their similarity.

\begin{figure*}
    \centering
    \includegraphics[width=.8\linewidth]{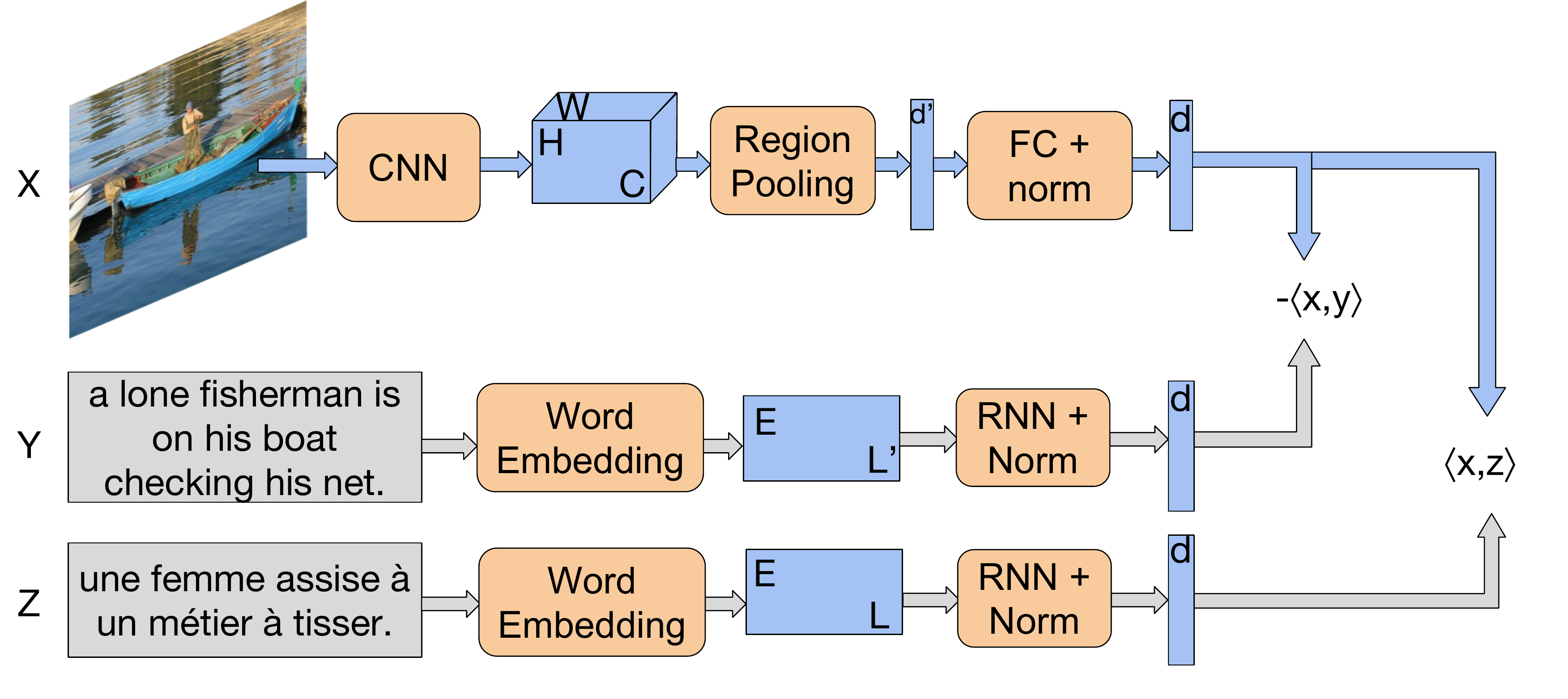}
    \caption{Proposed multilingual text image embedding architecture. This pipeline shows how during the training phase  how the model learn how to match an image with a sentence. $X$ and $Y$ are an image and a sentence that match each other. $Z$ is an unrelated sentence. For the loss computation, we want the dissimilarity between $X$ and $Y$ to be as big as possible and the dissimilarity between $X$ and $Z$ as small as possible.}
    \label{fig:full_model}
\end{figure*}

To do this, we train a triplet network~\cite{Hoffer2015, Schroff2015} that compares an image to two sentences: one that describes the image, another that does not (See figure~\ref{fig:full_model}).

\subsection{Overview}%
\label{sub:overview}


Our framework enables to take advantage of the availability of multilingual corpora in order to learn a cross-modal representation between texts and images.
We present a pipeline that learns a common representation between texts in different languages and images.
This common representation enables to compare image and texts using similarity measure for fast information retrieval.

The figure~\ref{fig:overview} illustrate the different step of the method to extract a representation from texts or images.
We show that we can learn a model in several languages and obtain state-of-the-art results on several retrieval tasks.
We also show that our method can improve state-of-the-art methods and generalize on languages the model have not seen during the training but which are model by \gls{muse}.
Our method is composed of two paths.

One path extract information from an image using a feature extractor like ResNet. Then a feature pooling method~\cite{Durand2016} enables to extract a signature from the features. We use the Weldon pooling method which automatically selects areas with highest and lowest activation and computes the image signature using these areas.

The second path computes an embedding of the input text using \gls{muse}. This method enables to compute close embeddings for words with close meanings in different languages. A \gls{rnn} is then used to extract a representation from the set of embeddings.

One interesting property of the Weldon pooling is that it produces a mapping between the highest response in the image path and the most significant words in the text path.
This property enables to extract an accurate representation between text and images, enabling to search precisely images with text and vice versa.

\subsection{Multilingual Sentence Embeddings}

To embed the sentence, our model first relies on individual word embeddings.
As we want to embed every sentence from every language in a unique vector space, we use word embeddings aligned in different languages.
The usual method consists in using \gls{bv}~\cite{Luong2015}.  \gls{bv} aligns two languages in the same space by learning the word embeddings on the two languages simultaneously.

There are 4 languages in our captions: English, French, German, Czech.
On the one hand, we propose to use pre-trained embeddings from \gls{muse}.  \Gls{muse} is a multilingual extension of FastText~\cite{Conneau2018} that embeds and aligns 30 languages in a single vector space. On the other hand, we propose to jointly use \gls{bv} and \gls{muse} approaches in order to enhance our multilingual representation.

The main idea is to train independently several bilingual word embeddings, in which, one of the languages is English.
Then, we learn a mapping between the different English representations (from the several bilingual word embeddings) to maximize the link between the bilingual representations.
For instance, one can train two bilingual word embedding models like English-French and English-Czech representations, apply the \gls{muse} approach on the two English parts.
From the two bilingual representation (English-French and English-Czech), we obtain a third one: French-Czech.

\begin{figure}
    \centering
    \includegraphics[width=\linewidth]{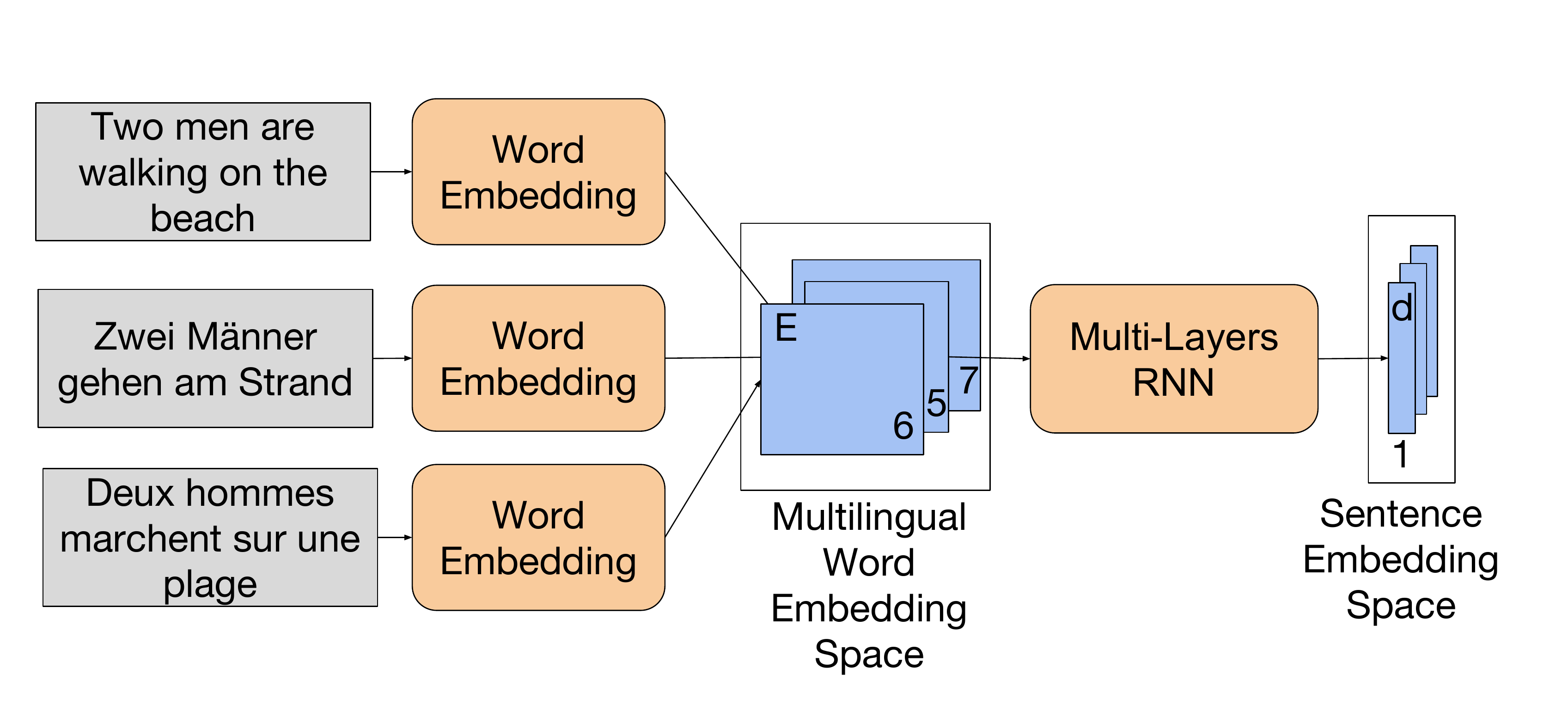}
    \caption{Sentences from different languages are mapped to a common Sentence embedding space. The word embedding method project words from different languages into a common space. For each word of the input sentence an embedding is computed and a \gls{rnn} is used to extract the embedding of the whole sentence.}
\label{fig:multilingualembedding}
\end{figure}

As shown on figure~\ref{fig:multilingualembedding}, we combine sentences from different languages by using word embedding models from different languages according to the approach described previously.

Those word embeddings being in the same space we can use a multi-layer \gls{rnn} to learn a sentence embedding.
This network is composed of 4 layers of \gls{sru}~\cite{Lei2017}, with a dropout after each layer.
The goal of this \gls{rnn} is to encode a vector of word embeddings of size $E$ into a Sentence Embedding Space $\mathbb{R}^d$.
Lastly, we normalize the output of the \gls{rnn} to obtain an embedding of the sentence in $[-1 , 1 ]^d$.

\subsection{Joint embedding}

The visual path of our network is similar to the one used by Engilberge et al.~\cite{Engilberge2018}.
It starts with a ResNet152~\cite{Xie2017}, on which we replaced its last fully connected layer (usually used for classification) with a Weldon pooling layer~\cite{Durand2016}. This layer pulls the regions with the maximum activation in the network, i.e. the regions of interest, and gives us an embedding of the image, a vector of dimension $d'$.
Finally, this vector goes through a fully connected layer that normalizes it, which aims to obtain an embedding of the image in $[-1 , 1 ]^d$.

Both pipelines are learned simultaneously, each image is paired with two sentences, one that describes the image, the other that doesn't.
The architecture of the model is shown on figure~\ref{fig:full_model}.
The two outputs are compared using a cosine similarity, which is equivalent to the inner product as both embeddings are normalized.

We use a triplet loss ~\cite{Wang2014, Schroff2015, Gordo2017a} to converge correctly and increase our performances.
This loss enables us to compare the relative similarity between the image and both sentences: the corresponding sentence should be closer to the image than the unrelated one.

The Figure~\ref{fig:full_model} presents the model with a triplet of one image and two captions.
The sentence $Y$ describes the image, and the caption $Z$ is an unrelated caption.
The triplet loss is shown in equation~\ref{eq:cost}, with $x$, $y$, and $z$ being respectively the embeddings of $X$, $Y$, and $Z$. $\alpha$ is the minimum margin between the similarity of the correct caption and the unrelated caption. During the training, it was set to $0.2$.

\begin{equation}
    \label{eq:cost}
    loss(x,y,z) = max(0, \alpha - x \cdot y + x \cdot z)
\end{equation}

\subsection{Training}

We train the ResNet on a classification task. This enables the \gls{cnn} to learn the extraction of interesting image features. We used a ResNet-152 pretrained on ImageNet ~\cite{Russakovsky2014} dataset, which provides a large collection of images, over 1 million, for 1000 categories.
The last layer of the ResNet, that was used for classification, is removed and replaced by a Weldon pooling, followed by a randomly initialized fully connected layer with a dropout regularization.

For the text pipeline, we use pre-trained \gls{w2v}, FastText, \gls{bv} and \gls{muse} word embeddings.
We then freeze the \gls{cnn} and region pooling of the network and train the \gls{rnn} and the fully connected layer. This enables to project the embeddings of both sentence and image into a common space.
Finally, we fine-tune the entire network.

As shown by Schroff et al.~\cite{Schroff2015}, the triplets used to train the model should be carefully selected.
Indeed, by using \textquote{easy} triplets, i.e. triplets on which the network performs already well, the network learns almost nothing.
As proposed previously in ~\cite{Faghri2018, Engilberge2018}, we aim to focus on the \textquote{hardest} triplets only, i.e. the hardest ones to differentiate by the network.
Instead of looking for the best triplet throughout the entire dataset at each iteration, we stay in the current batch.
For each image and its corresponding sentence $(i,s)$, we select the closest non-similar image to $s$ and the closest non-similar sentence to $i$.

The following equation shows the loss computation over the batch $B$ composed of image and caption pairs $(i,s)$:

\begin{equation}
    \sum_{i,s \in B} \left (  \max_{z \in U_i} loss(i,s,z) + \max_{z \in D_i} loss(s,i,z)       \right )
\end{equation}

Where $D_i$ represents every image in the batch $B$ that differs from $i$.
$U_i$ represents every sentence unrelated to the image $i$, inside the batch $B$, in every language.
Each batch can contain several captions in multiple languages corresponding to the same image.
This enables the selection of the best example inside each training batch.

\subsection{Visualization}

To enhance the recall evaluation made previously, we propose some visual evaluation.
The figure~\ref{fig:closestImages} shows the five closest images for the same sentence in French and German.
These images come from the Google Semantic Caption dataset, which contains \num{3000000} images.

\begin{figure*}
    \centering
    \begin{subfigure}{\textwidth}
        \centering
        \foreach \x in {1, ..., 5}{
            \begin{subfigure}{.19\textwidth}
                \centering
                \includegraphics[height=2cm]{figures/violin_de_\x.jpg}
                \label{fig:violin_de_\x}
            \end{subfigure}
        }
        \caption{\textquote{eine Frau Geige spielt}\label{fig:violin_de}}
    \end{subfigure}
    \begin{subfigure}{\textwidth}
        \centering
        \foreach \x in {1, ..., 5}{
            \begin{subfigure}{.19\textwidth}
                \centering
                \includegraphics[height=2cm]{figures/violin_fr_\x.jpg}
                \label{fig:violin_fr_\x}
            \end{subfigure}
        }
        \caption{\textquote{Une femme jouant du violon}\label{fig:violin_fr}}
    \end{subfigure}
    \begin{subfigure}{\textwidth}
        \centering
        \foreach \x in {1, ..., 5}{
            \begin{subfigure}{.19\textwidth}
                \centering
                \includegraphics[height=2cm]{figures/violin_en_\x.jpg}
                \label{fig:violin_en_\x}
            \end{subfigure}
        }
        \caption{\textquote{A woman playing violon}\label{fig:violin_en}}
    \end{subfigure}
    \caption{Closest images for the same sentence in different languages from the Google Semantic Caption dataset. Although the sentence is the same in 3 different languages, we can see that the results are slightly different. This is explained by the words embeddings which are close between languages but not the same.}
    \label{fig:closestImages}
\end{figure*}

\begin{figure}
    \centering
    \begin{subfigure}{.24\columnwidth}
        \centering
        \includegraphics[height=2.cm]{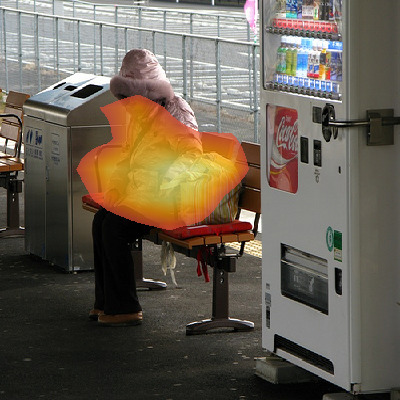}
        \caption*{woman}
    \end{subfigure}
    \begin{subfigure}{.24\columnwidth}
        \centering
        \includegraphics[height=2.cm]{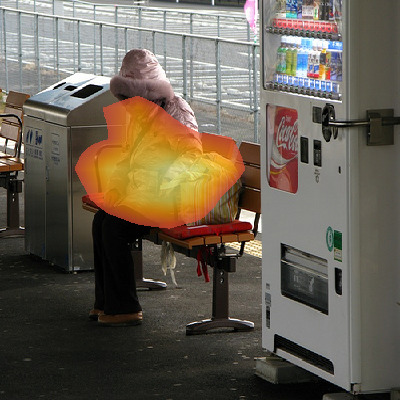}
        \caption*{femme}
    \end{subfigure}
    \begin{subfigure}{.24\columnwidth}
        \centering
        \includegraphics[height=2.cm]{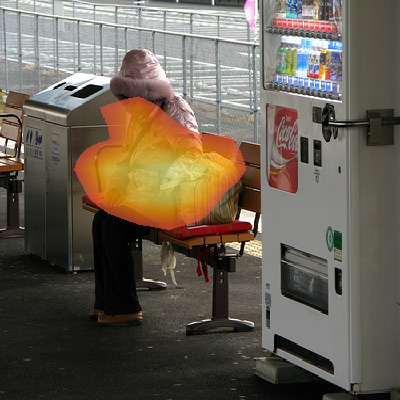}
        \caption*{Frau}
    \end{subfigure}
    \begin{subfigure}{.24\columnwidth}
        \centering
        \includegraphics[height=2.cm]{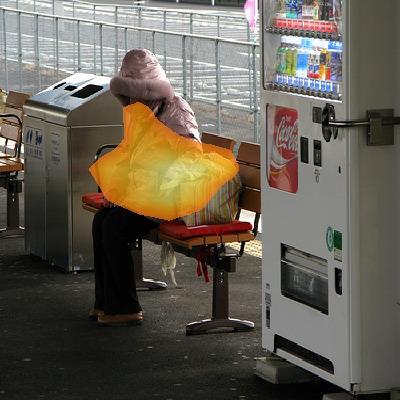}
        \caption*{žena}
    \end{subfigure}

    \begin{subfigure}{.24\columnwidth}
        \centering
        \includegraphics[height=2cm]{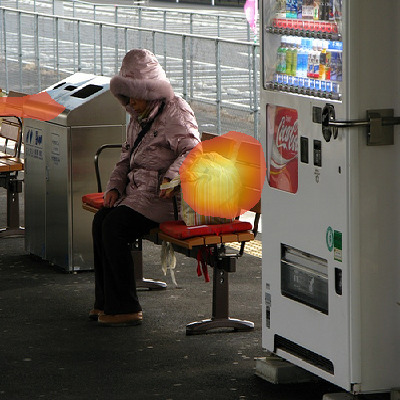}
        \caption*{bag}
    \end{subfigure}
    \begin{subfigure}{.24\columnwidth}
        \centering
        \includegraphics[height=2.cm]{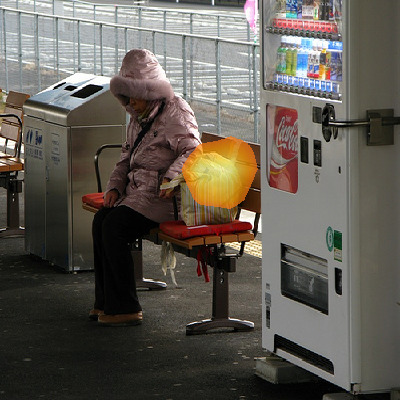}
        \caption*{sac}
    \end{subfigure}
    \begin{subfigure}{.24\columnwidth}
        \centering
        \includegraphics[height=2.cm]{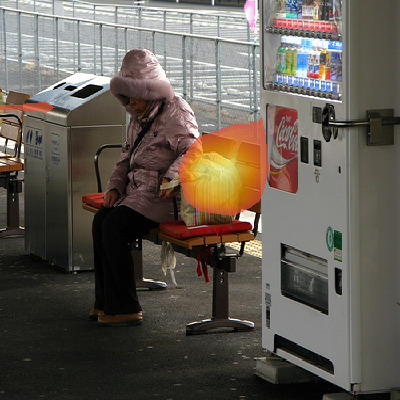}
        \caption*{Beutel}
    \end{subfigure}
    \begin{subfigure}{.24\columnwidth}
        \centering
        \includegraphics[height=2.cm]{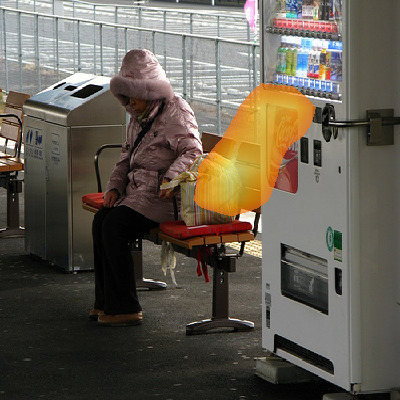}
        \caption*{sáček}
    \end{subfigure}

    \begin{subfigure}{.24\columnwidth}
        \centering
        \includegraphics[height=2.cm]{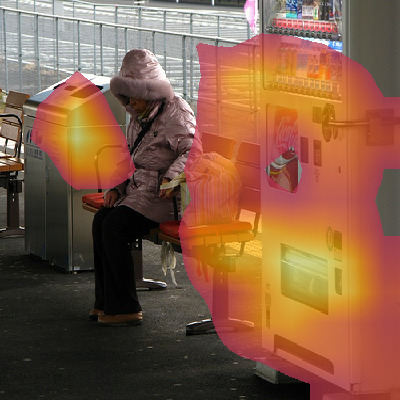}
        \caption*{machine}
    \end{subfigure}
    \begin{subfigure}{.24\columnwidth}
        \centering
        \includegraphics[height=2.cm]{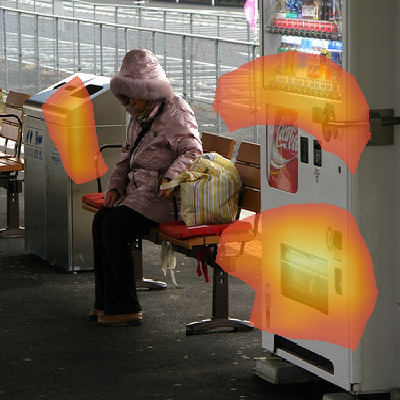}
        \caption*{machine}
    \end{subfigure}
    \begin{subfigure}{.24\columnwidth}
        \centering
        \includegraphics[height=2.cm]{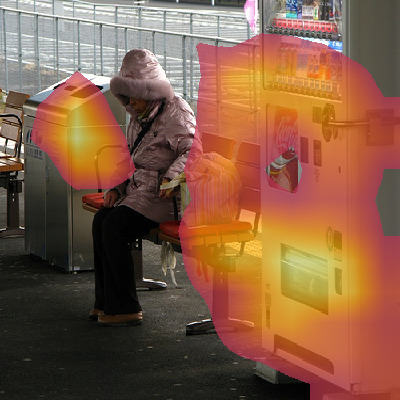}
        \caption*{Maschine}
    \end{subfigure}
    \begin{subfigure}{.24\columnwidth}
        \centering
        \includegraphics[height=2.cm]{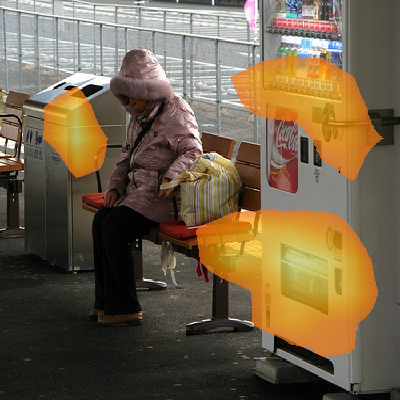}
        \caption*{stroj}
    \end{subfigure}

    \begin{subfigure}{.24\columnwidth}
        \centering
        \includegraphics[height=2.cm]{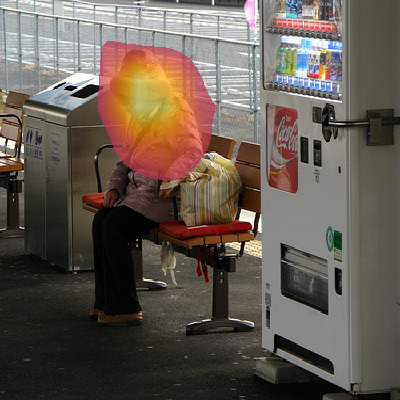}
        \caption*{jacket}
    \end{subfigure}
    \begin{subfigure}{.24\columnwidth}
        \centering
        \includegraphics[height=2.cm]{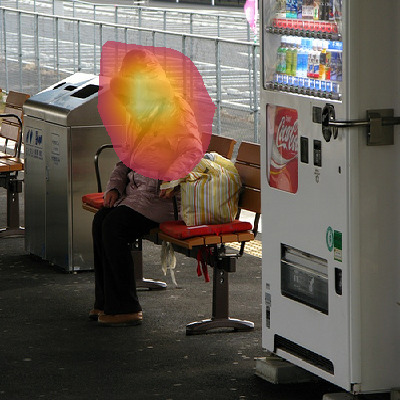}
        \caption*{manteau}
    \end{subfigure}
    \begin{subfigure}{.24\columnwidth}
        \centering
        \includegraphics[height=2.cm]{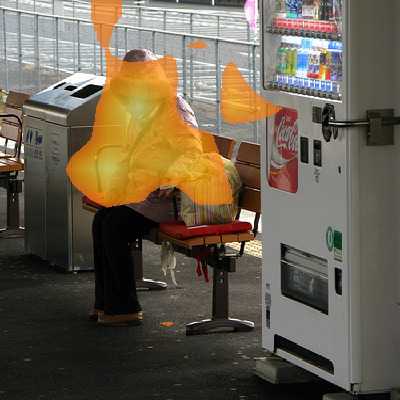}
        \caption*{Mantel}
    \end{subfigure}
    \begin{subfigure}{.24\columnwidth}
        \centering
        \includegraphics[height=2.cm]{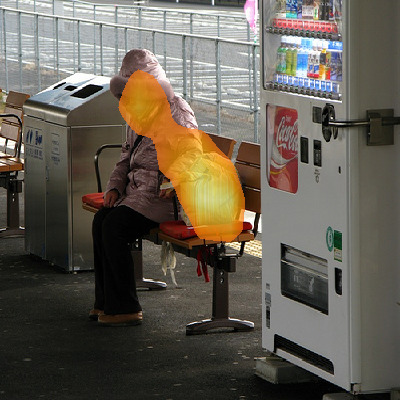}
        \caption*{plášť}
    \end{subfigure}
    \caption{Activation maps of the network with different input words in different languages. The first column shows the activation map for English, the second for French, the third for German and the last for Czech. Theses activations show the ability of the network to recognizing the important areas of the image according to the input word. This confirms that the method is able to match an object in the image with an associated word.}
\label{fig:heatmap}
\end{figure}

We show, in Figure~\ref{fig:heatmap}, the maximum of activation of the network, given different words.

We observe the activation zone of the \gls{cnn} depending on the word used for the \gls{rnn}.
The network responds mostly in the same way for words from different languages.
Some differences appear with less common words like \textquote{Mantel} in German, with a noisier activation than for the French\textquote{Manteau} or the English \textquote{Jacket}.


\section{Experiments}\label{sec:xp}  

In this section, we present our experimental protocol: hardware/software setup, datasets used and numerical results. We evaluate and compare our method with state-of-the-art approaches by using classic metrics.

\subsection{Experimental setup}

All the experiments were done on an NVIDIA {DGX-1}\footnote{\url{https://www.nvidia.com/}}.
Training the network on \gls{coco} and Multi30k take roughly two days on 4 V100 GPUs with 16GB of RAM, for each experiment.
We used Facebook implementation of FastText\footnote{\url{https://fasttext.cc/}} to compute word embeddings.
Finally, we rely on Pytorch\footnote{\url{https://pytorch.org/}} for deep learning implementation.

\subsection{Datasets}

To train and evaluate our model, we used three datasets of images with their corresponding captions.
The first dataset is \gls{coco} ~\cite{Lin2014}. It contains \num{123287} images with 5 English captions per image. We used the \textit{val split} from Karpathy et al.~\cite{Karpathy2015} (\num{113287} train, \num{5000} validation and \num{5000} test images) to train and evaluate our model on English sentences.

To train our model on other languages, we used the Multi30K~\cite{Elliott2016} dataset, containing \num{31014} images with captions in French, German, and Czech. \num{29000} are kept for training, \num{1014} for validation and \num{1000} for testing.
Lastly, for evaluation purposes we used Google's Conceptual Captions\footnote{\url{https://ai.google.com/research/ConceptualCaptions}} dataset, containing \num{3154240} captioned images.

\subsection{Evaluation method}

We evaluate the quality of our results using $recall@k$, which is
the proportion of relevant images found in the top-$k$ returned images for a given query.
We evaluate caption retrieval with images as queries using recall at $1$, $5$ and $10$.
This means that we verify if the sentence corresponding to an image is in the first, fifth, or tenth closest results.
For image retrieval, each caption is evaluated in the same way. The presence of the image corresponding to a sentence is verified in the first, fifth, or tenth closest results.

The caption retrieval test is made in batches of 1000 images and caption pairs, using the \gls{coco} dataset.
In the Multi30K dataset, every images have a caption in each language. The recall is computed across languages.



\subsection{Results and analysis}

We perform four different experiments. 
The aim of the first and second experiments is to verify the performance of our model with English captions, depending on the word embeddings used.
They measure respectively the caption and the image recall in English on \gls{coco} dataset.

Finally, in experiment 3 and 4, we evaluate the model with caption and image recall, on the Multi30K dataset, with captions in different languages.

\paragraph{Experiment 1.}

The models are trained on the \gls{coco} dataset for English and on Multi30K dataset for French, German and Czech.
We use the \gls{coco} dataset for evaluation.

\begin{table}%
    \centering
    \caption{Experiment 1: Caption retrieval on the \gls{coco} dataset. We compare the different reminders of the different methods first in English and then by adding new languages. We also evaluate variations of \acrshort{dsve} method with different word embedding.\label{tab:recall_coco}}
    \rowcolors{1}{white}{gray!15}
    \begin{tabular}{llSSS}
        \toprule
        Embedding                                     & lang.    & \text{r@1}      & \text{r@5}      & \text{r@10} \\
        \midrule
        \acrshort{vse}~\cite{Faghri2018}              & en       & 64.6            & $\emptyset$        & 95.7 \\
        \acrshort{dsve}~\cite{Engilberge2018}         & en       & \bfseries 69.8  & \bfseries 91.9  & 96.6 \\
        \hline
        \acrshort{dsve} w/ \acrshort{w2v} & en       & 63.48           & 89.48           & 95.64 \\
        \acrshort{dsve} w/ FastText       & en       & 66.08           & 90.7            & 96.2 \\
        \hline
        Ours  w/ \acrshort{bv}             & en       & 65.58           & 90.52           & 96.1 \\
                                                      & en+fr    & \bfseries 67.78 & \bfseries 91.58 & \bfseries 96.92 \\
        Ours w/ \acrshort{muse}           & en       & 63.1            & 89.58           & 95.56 \\
                                                      & en+fr    & 63.88           & 89.2            & 95.24 \\
                                                      & en+fr+de & 62.4            & 89.18           & 95.16 \\
                                                      & all      & 63.28           & 88.3            & 94.6 \\
        \bottomrule
    \end{tabular}
\end{table}

The table~\ref{tab:recall_coco} shows the caption retrieval recall on \gls{coco} dataset.
The first two lines show the state-of-the-art results.
The second pair of lines present the results of our model, with \gls{w2v} and FastText embeddings used as the baseline.
We can see that our model is close to the \gls{dsve} method~\cite{Engilberge2018} while the \gls{w2v} method is slightly worst, as the representation power of the word embedding is reduced.

The \gls{bv} English-French method is used in English and on both languages simultaneously.
If trained only on English, i.e. only on the \gls{coco} dataset like the two previous methods, it shows performance similar to the one of the state-of-the-art.
This means training using \gls{bv} does not weaken the English representation. 
When trained on English and French together, the recall is increased by \SI{3.35}{\percent}, going from \SI{65.58}{\percent} to \SI{67.78}{\percent}.
We can also see an improvement for recall@5 and recall@10, with respectively \SI{1.17}{\percent} and \SI{0.85}{\percent} of increase.
This implies that the similarity learning with French captions increases English recognition when using \gls{bv}.

To verify if we can generalize this result to a larger number of languages, we used \gls{muse}  aligned for 30 languages.
Using the Multi30K dataset, we can also train the model on German (de) and Czech (cs).

First of all, when training with \gls{muse} for English only, we can see a sharp decrease in performance, with a recall going from \SI{66.08}{\percent} to \SI{63.1}{\percent}.
By comparing the model trained with \gls{w2v}, we obtain similar results.
This could come from the fact that both \gls{muse} and \gls{w2v} embeddings do not have representation for out of vocabulary words like the FastText ones.
Moreover, rare words have much more chance to be wrongly projected because of space transformation.
When we train the model with additional languages, we can see a slight decrease in performance in English.
The maximum decrease is \SI{1.01}{\percent} for recall@10, but it is counter-balanced by an increase of \SI{0.29}{\percent} for the recall@1.

\paragraph{Experiment 2.}

Given a sentence, in any language, we evaluate the rank of the corresponding image.
The evaluation is again made by batches of \num{1000}.
The results are presented in table~\ref{tab:recall_cocoimage}.

The first two lines of the table present the state-of-the-art results, with \gls{w2v} and FastText embeddings.
We can see similar results as in the previous experiment.
With \gls{bv}, we have results close to the FastText embeddings when training only in English.
This time, the recall is better with an increase of \SI{2.68}{\percent} for recall@1.
When trained with English and French, the recall@1 is increased by \SI{3.65}{\percent}.
This implies, again, that we can improve performance by learning on an additional language.

Our model is able to use the multi-language representing the effectiveness of \gls{muse} embeddings.
We train the model with English, and different combinations of French, German and Czech.
On English only, we have similar results to the \gls{w2v} approach.
When adding new languages, we can see a decrease in performance for English.
We obtain a maximum decrease of \SI{2.62}{\percent} for recall@1 when the models saw English, French, German and Czech.

\begin{table}%
\centering
\caption{Experiment 2: Image retrieval on the \gls{coco} dataset. The methods are the same as in table~\ref{tab:recall_coco}.\label{tab:recall_cocoimage}}
 \rowcolors{1}{white}{gray!15}
 \begin{tabular}{llSSS}
    \toprule
    Embedding                             & lang.    & \text{r@1}      & \text{r@5}      & \text{r@10}\\
    \midrule
    \acrshort{vse}~\cite{Faghri2018}      & en       & 52.00           & $\emptyset$        & 92.0 \\
    \acrshort{dsve}~\cite{Engilberge2018} & en       & 55.9            & 86.9            & 94.0 \\
    \hline
    \acrshort{dsve} w/ \acrshort{w2v}     & en       & 51.87           & 84.31           & 92.48 \\
    \acrshort{dsve} w/ FastText           & en       & 54.12           & 85.74           & 92.93 \\
    \hline
    Ours w/ \acrshort{bv}                 & en       & 55.57           & 86.92           & 93.86 \\
                                          & en+fr    & \bfseries 56.09 & \bfseries 87.22 & \bfseries 94.03 \\
    Ours w/ \acrshort{muse}               & en       & 51.81           & 84.70           & 92.82 \\
                                          & en+fr    & 52.25           & 84.72           & 92.74 \\
                                          & en+fr+de & 51.17           & 84.09           & 92.22 \\
                                          & all      & 50.44           & 83.39           & 91.80 \\
    \hline
 \end{tabular}
\end{table}

\paragraph{Experiment 3.}

The model is trained with English only, then with English and French (en+fr), with English, French, and German (en+fr+de) and with English, French, German and Czech (all).
We can see a decrease in performance when adding French that is not present with other languages.
Otherwise, every time we add a new language the recall for this language logically increase.
The best performance is achieved with English+French+German+Czech, with an increase of \SI{6.42}{\percent} for multilingual retrieval.

\begin{table}%
\centering
 \caption{Image Recall@10 on the Multi30k dataset with different languages with \gls{muse}.\label{tab:multi30k_matrix_muse}}
 \rowcolors{1}{white}{gray!15}
 \begin{tabular}{lSSSSS}
    \toprule
    train. lang. & \text{en}       & \text{fr}       & \text{de}       & \text{cs}       & \text{all} \\
    \midrule
    en           & \bfseries 56.60 & 46.05           & 44.18           & 38.75           & 46.40 \\
    en+fr        & 50.93           & 43.69           & 41.61           & 34.02           & 42.43 \\
    en+fr+de     & 54.63           & 46.94           & 45.07           & 38.26           & 46.22 \\
    all          & 55.32           & \bfseries 49.30 & \bfseries 46.84 & \bfseries 46.06 & \bfseries 49.38 \\
    \bottomrule
 \end{tabular}
\end{table}

\paragraph{Experiment 4:}

With \gls{bv} embeddings, we learn two languages at the same time, and test retrieval on one or two of these languages.
Results are shown in table~\ref{tab:multi30k_matrix_BIVEC}.
Trained in English alone, the model gives worse performance than \gls{muse} for languages not seen previously.
For example, with English-German \gls{bv} and a model trained only in English, and test on German, we obtain only \SI{22.96}{\percent} recall@10, where \gls{muse} embeddings obtain \SI{44.18}{\percent}.
But when train on English and French, we obtain \SI{55.22}{\percent} recall, an increase of \SI{26.39}{\percent} compared to \gls{muse}.
With German and English training, we have an increase of \SI{15.16}{\percent} on English only recall, with a recall of \SI{61.44}{\percent}.
Meaning that, once again, learning a new language with \gls{bv} enables better results in English, as the same kind of results is visible with French as well.

\begin{table}%
\centering
 \caption{Image Recall@10 on Multi30k dataset with different languages with \gls{bv} Embeddings.\label{tab:multi30k_matrix_BIVEC}}

 \rowcolors{1}{white}{gray!15}
 \begin{tabular}{lSSSSS}
    \toprule
    train. lang. & \text{en}       & \text{fr}       & \text{de}       & \text{en+fr}    & \text{en+de}\\
    \midrule
    en           & 53.35           & 26.13           & 22.96           & 39.74           & 34.57 \\
    en+fr        & 59.76           & \bfseries 55.22 & $\emptyset$     & \bfseries 57.50 & $\emptyset$\\
    en+de        & \bfseries 61.44 & $\emptyset$     & \bfseries 43.59 & $\emptyset$     & \bfseries 52.51 \\
    \bottomrule
 \end{tabular}
\end{table}

\section{Conclusion}
\label{sec:conclusion}

We presented a novel approach for multilingual text and image embeddings.
While the method provides close to the state-of-the-art performance in English-only context, its main advantage is that it enables the use of multilingual dataset in order to improve the performance.
We showed that using a new language during the training process improve image retrieval for other languages.

Our method uses a \gls{cnn} to extract image information. It also uses aligned multilingual word embeddings and a \gls{rnn} to produce text representations.
This way, it provides image and text embeddings that can be trained to produce comparable features.
We demonstrate that we can use this network to retrieve an image from a text and text from an image, with a multilingual representation.

We evaluated our method on the \gls{coco} dataset for English-only results and shown that using \gls{bv} embeddings enables the use of another language in order to improve the performance.
The obtained improvement is a \SI{3.35}{\percent} increase in performance on the \gls{coco} dataset, and a \SI{15.15}{\percent} increase on the Multi30K dataset.
By using \gls{muse} embeddings, we are able to embed more languages in the same model.
We showed that adding other languages decrease performance for English, but increase the recall in a multilingual environment.
For image retrieval from a caption in 4 languages, we obtain a \SI{49.38}{\percent} recall@10 on the Multi30K dataset.


{\small
\bibliographystyle{ieee}
\bibliography{references.bib}
}
\newpage

\end{document}